\documentclass[10pt,twocolumn,letterpaper]{article}

\usepackage{cvpr}              %

\usepackage{fontawesome}

\DeclareMathOperator{\mlimm}{LIM}
\DeclareMathOperator{\mlrm}{LRM}
\DeclareMathOperator{\mlrmmv}{LRM}  %
\newcommand{\src}{\text{src}}
\newcommand{\tgt}{\text{tgt}}

\newcommand{\limm}{$\mlimm$\xspace}
\newcommand{\lrm}{$\mlrm$\xspace}
\newcommand{\lrmmv}{$\mlrmmv$\xspace}

\newcommand{\webpagelink}[1]{\href{https://remysabathier.github.io/lim.github.io/}{#1}}
\newcommand{\webpagelinkflat}[1]{\href{https://remysabathier.github.io/lim.github.io/}{#1} [https://remysabathier.github.io/lim.github.io]}

\makeatletter
\renewcommand{\paragraph}{%
    \@startsection{paragraph}{4}%
    {\z@}{-0.5em}{-0.5em}%
    {\normalfont\normalsize\bfseries}%
}
\makeatother

\definecolor{cvprblue}{rgb}{0.21,0.49,0.74}
\usepackage[pagebackref,breaklinks,colorlinks,allcolors=cvprblue]{hyperref}

\title{LIM: Large Interpolator Model for Dynamic Reconstruction}

\author{
Remy Sabathier\\
University College London and Meta
\and
Niloy J. Mitra\\
University College London
\and
David Novotny\\
Meta
}

\begin{document}

\maketitle

\begin{abstract}

Reconstructing dynamic assets from video data is central to many in computer vision and graphics tasks. Existing 4D reconstruction approaches are limited by category-specific models or slow optimization-based methods. Inspired by the recent Large Reconstruction Model~(LRM)~\cite{hong_lrm_2023}, we present the Large Interpolation Model~(LIM), a transformer-based feed-forward solution, guided by a novel causal consistency loss, for interpolating implicit 3D representations across time. 
Given implicit 3D representations at times $t_0$ and $t_1$, LIM produces a deformed shape at any continuous time $t\in[t_0,t_1]$, delivering high-quality interpolated frames in seconds.
Furthermore, LIM allows explicit mesh tracking across time, producing a consistently uv-textured mesh sequence ready for integration into existing production pipelines. We also use LIM, in conjunction with a diffusion-based multiview generator, to produce dynamic 4D reconstructions from monocular videos. We evaluate LIM on various dynamic datasets, benchmarking against image-space interpolation methods (e.g., FiLM~\cite{reda2022filmframeinterpolationlarge}) and direct triplane linear interpolation, and demonstrate clear advantages. In summary, LIM is the first feed-forward model capable of high-speed tracked 4D asset reconstruction across diverse categories. Video results and code are available via the \href{https://remysabathier.github.io/lim.github.io/}{project page}.

\end{abstract}
    
\section{Introduction}
\label{sec:intro}

Reconstructing dynamic 4D assets from video data is a fundamental problem in computer vision and graphics, with many virtual and augmented reality applications.
Existing 4D reconstructors follow two main paradigms: category-specific articulated reconstruction and image-or-text conditioned 4D distillation.
Hence, they are either restricted to a specific class of objects such as humans \cite{loper15smpl:} and animals \cite{biggs18creatures,animalAvatars:24}, or are optimization-based \cite{yin1234dgen:,ren_dreamgaussian4d_2023} making them slow, requiring minutes to hours per reconstruction.

\begin{figure}[t!]
    \centering
    \includegraphics[width=\columnwidth]{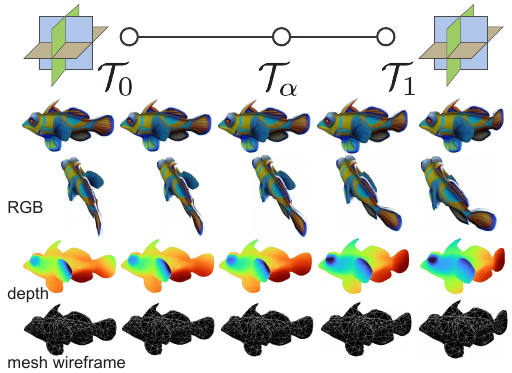}
    \caption{
    \textbf{Large Interpolator Model (LIM)}
    outputs a 4D video reconstruction by interpolating 3D implicit representations between two consecutive keyframes at times $t=0$ and $t=1$, which can then be used to produce 3D-consistent RGB, depth, or decoded as tracked mesh sequences. 
    }
    \label{fig:teaser}
\end{figure}

Recently, in the context of static reconstruction, the large reconstruction model~(LRM)~\cite{hong24lrm:} has been proposed as an elegant feed-forward network that, starting from a fixed rig of multiview images, 
directly produces 3D implicit representation, which can then be rendered for novel view generation.
In this work, in the context of dynamic reconstruction, we ask \textit{if a similar feed-forward approach can be developed to reconstruct a tracked explicit representation across time}.

Here, L4GM~\cite{ren2024l4gm} proposed a feedforward 4D video reconstructor which, for each video keyframe, accepts few views of the reconstructed object and outputs a mixture of 3D Gaussian Splats \cite{kerbl233d-gaussian}.
However, this approach has limitations as it can only reconstruct the keyframes at their exact timesteps without the ability to interpolate the shape through time.
Additionally, establishing correspondences between Gaussian mixtures from different timesteps is challenging, which complicates tracing the deformation of the underlying object geometry through time.
This limitation hinders many important downstream applications, such as gaming, where we require tracked meshes in the form of the 3D shape and texture of a single mesh to be defined in a static canonical pose, with only its geometry (i.e., vertices) allowed to be deformed across time. 

We thus present Large Interpolation Model~(LIM) as a transformer-based feed-forward solution that accepts an implicit representation of an object at two different keyframe times $t_0$ and $t_1$ of a video, and interpolates between the two at any continuous intermediate timestep $t \in [t_0, t_1]$.
We enable this with a novel self-supervised \textit{causal consistency loss} that allows us to meaningfully interpolate continuously in time, even when supervised with keyframes from distinct time stamps. 
LIM is not only an efficient interpolator, but can also track a source mesh across time producing a functional deformable 3D asset with a shared uv texture map.
Here, LIM tracks the mesh by means of an additional volumetric function that maps each time-specific 3D implicit-surface point to a unique coordinate on the intrinsic (time-invariant) surface of the object.
This is unique -- unlike any other competing dynamic reconstructor \cite{ren2024l4gm}, LIM outputs a mesh with time-invariant texture and topology, and time-dependent vertex deformation. 
This renders LIM directly applicable in existing production setups.

Our LIM module also enables dynamic reconstruction from monocular video.
Specifically, given keyframes of a monocular video, a pretrained image diffusion model generates additional object views which, using a multiview LRM, we convert to keyframe-specific implicit 3D representations.
Then, LIM directly interpolates the 3D representations yielding a dynamic 4D asset.

Our experiments demonstrate that LIM outperforms existing alternatives in terms of the overall quality of the implicit-shape interpolations while being several times faster.
Furthermore, we also evaluate the quality of the mesh tracing, where LIM records significant performance improvements.

\begin{figure*}[t]
\centering
\includegraphics[width=\textwidth]{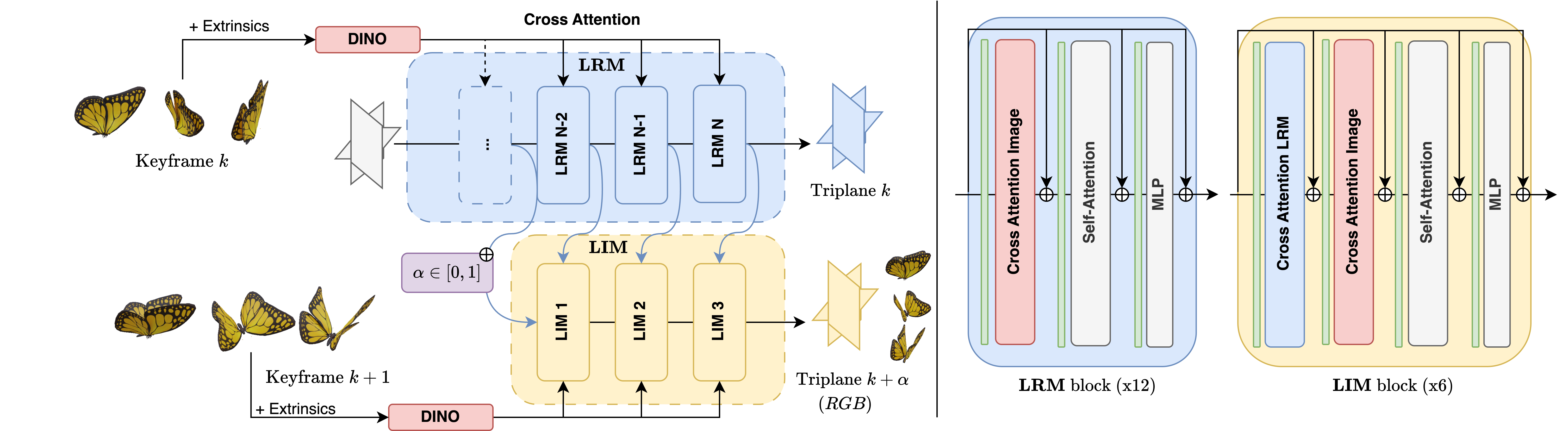}
\caption{\textbf{LIM framework}. (Left) Given multi-view images on 2 timesteps $k$ and $k+1$, \limm interpolates any intermediate 3D representation at $k+\alpha, \alpha \in [0,1]$. It achieves this notably via cross-attention with the latest intermediate features of \lrmmv on keyframe $k$. In practice, our \limm architecture has 6 blocks and \lrmmv 12 blocks. (Right) Block structure of \lrmmv and \limm. We include layer normalization before each module in blocks.}
\label{fig:lim_architecture}
\end{figure*}

\section{Related Work}
\label{sec:related_work}

\paragraph{3D Reconstruction.}
Early work, introduced by DreamFusion \citep{poole23dreamfusion:} optimizes a 3D scene via \emph{score distillation sampling} from a pretrained text-to-image diffusion model \citep{melas-kyriazi23realfusion, qian23magic123:, tang23make-it-3d:}.
However, these methods are slow to optimize and suffer from inconsistencies (like the Janus problem). 
Zero123 \citep{shi23zero123:} learns to condition diffusion models on a single-view image and camera transformation, which allows novel view generation.
Multiple novel views of a single object can then be used to optimize a NeRF reconstruction which, however, is often impaired by view-inconsistencies of the novel-view generator.
SyncDreamer \citep{liu23syncdreamer:} proposes an extension that improves the consistency of novel views and transitivelly of the generated 3D shapes.
Due to the highly-challenging task of reconstructing any 3D asset from a single image, several works \citep{yang_banmo_2023, wu23magicpony, wu_dove_2022, goel_shape_2020, kato_neural_2017, kanazawa_learning_2018, wang_pixel2mesh_2018, henderson_learning_2019, kato_learning_2019, li_online_2020, li_online_2020, li20self-supervised, kokkinos_learning_2021, kokkinos21to-the-point:} learn 3D reconstructors of a specific category which simplifies learning of shape, deformation, and appearance priors.
We also note some works on learning a generalizable dynamic radiance field from monocular videos \citep{sinha22common, tian_mononerf_2023} which, however, are not designed for outputting a time-deforming 3D mesh.
Recent methods \citep{hong_lrm_2023,wei24meshlrm:}, trained on large 3D datasets such as 
\emph{Objaverse} \citep{deitke23objaverse:, deitke23objaverse-xl:}, propose feed-forward reconstructors which directly predict 3D representation of an object, conditioning on a single or multiple views.
These methods dramatically reduce reconstruction speed as they don't rely on any optimization loop.

\paragraph{4D Representations.}
Extending the popular research on representing static 3D scenes with implicit shapes, recent works proposed new time-deforming alternatives.
\emph{Dynerf} \citep{gao_dynamic_2021} extends static neural radiance field \citep{mildenhall20nerf:} with an additional compact latent code to represent time deformation.
However, similar to the static implicit shape reconstructors \cite{mildenhall20nerf:,yariv21volume}, its optimization process is relatively slow.
\citep{cao23hexplane:, fridovich-keil23k-planes:, chen22tensorf:} factorize a dynamic representation into multiple low-rank components, which dramatically speeds up the optimization.
Notably, \emph{Hexplane} \citep{cao23hexplane:} proposes a 6-plane representation which extends the spacial triplane representation \cite{chan22efficient} to a spatio-temporal one.
With the emergence of \emph{3D Gaussian Splatting} \citep{kerbl_3d_2023} (3DGS),
\citep{wu_4d_2023, luiten_dynamic_2023} propose its extension to dynamic scenes, relying either on a per-frame optimization with dynamic constraints, or on a temporal network to deform the gaussians in time.

\paragraph{4D Generation and Reconstruction.}
Several works focus on text-to-4D generation: MAV3D \citep{singer_text--4d_2023} optimizes a Hexplane \citep{cao23hexplane:} representation via score distillation sampling from a text-to-image and a text-to-video diffusion model.
4Dify \citep{bahmani_4d-fy_2023} introduces a 3D-aware text-to-image diffusion model, and parameterizes the representation with a multi-resolution hash encoding \citep{muller22instant}.
However, these methods tend to produce very limited and simple motions.
TC4D \citep{bahmani_tc4d_2024} proposes an extension to decompose movement into local deformation and global rigid motion.
\citep{ling_align_2024} applies same SDS supervision with Gaussian splatting.

Similar to us, recent work focused on video-to-4D reconstruction.
Consistent4D \citep{jiang_consistent4d_2023} generates 4D content from monocular video via SDS supervision, optimizing a Cascade DyNerf \citep{gao_dynamic_2021}.
Simiarly, 4DGen \citep{yin_4dgen_2024} and DreamGaussian4D \citep{ren_dreamgaussian4d_2023} encode the 4D asset as a set of static 3D gaussians and a regularized deformation field.
\citep{pan_fast_2024, zeng_stag4d_2024} leverages diffusion models to generate frames across views and timestamps, and optimizes a dynamic gaussian splats based on these frames.
\citep{huang_sc-gs_2024, wu_sc4d_2024} decompose motion and appearance in gaussian splatting: instead of predicting a deformation for each gaussian in a canonical frame, they deform Gaussians by means of sparse control points.
All above methods are relatively slow due to the 2nd reconstruction stage that optimizes each 4D asset from scratch.
Furthermore, these methods cannot easily trace the resulting 4D asset through time, which prohibits their application in production setups.

\section{Method}
\label{sec:method}

In \cref{sec:method_lrm}, we review the LRM \citep{hong_lrm_2023} architecture which our method is based on; in \cref{sec:method_lim}, we introduce \limm, our large interpolator model, for efficient 3D interpolation and; in \cref{sec:trace_shapes}, we show how \limm can be used for fast 4D reconstruction and mesh tracking.

\begin{figure}[b!]
\centering
\includegraphics[width=\linewidth]{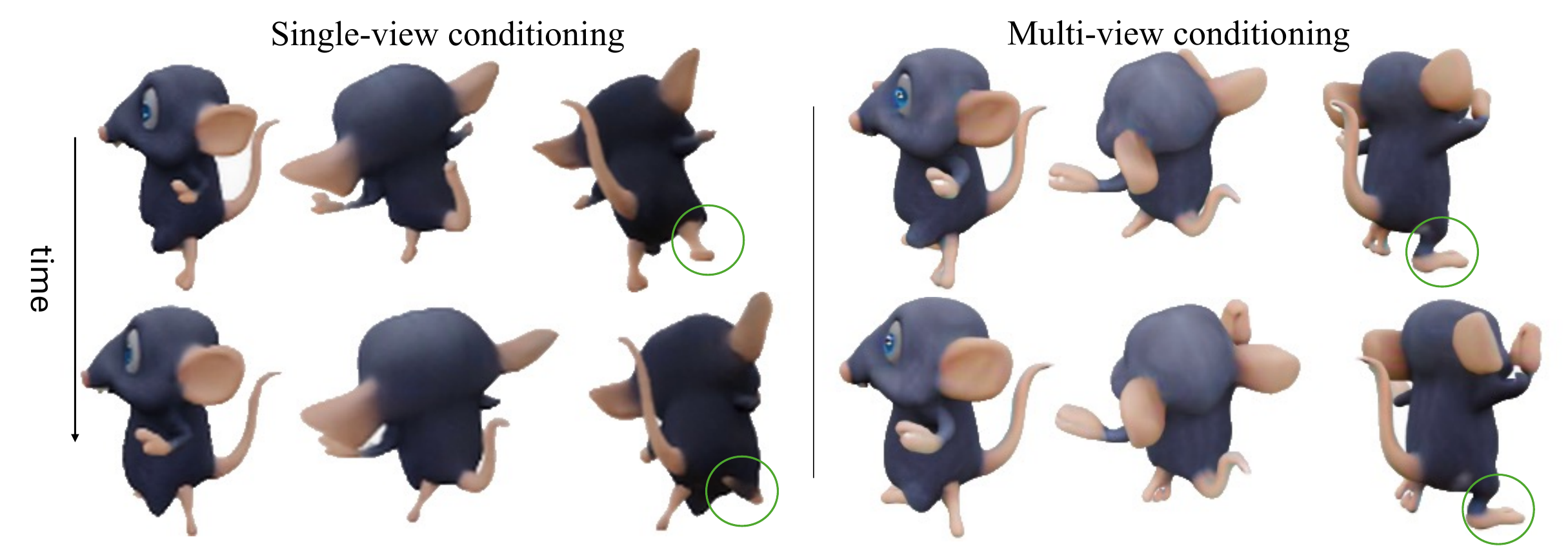}
\caption{LRM conditioned on a single-view \citep{tochilkin_triposr_2024} is sensitive to small changes on the input image, which gives inconsistent result from one video frame to another. 
The multi-view LRM prevents this instability. For each model, left shows an input-view, right shows two target views. Each line is a different timestep.}
\vspace{-0.1in}
\label{fig:inconsistentview}
\end{figure}

\subsection{Preliminaries} \label{sec:method_lrm}

Our Large Interpolation Model (\limm) is built upon the multi-view version of Large Reconstruction Model (\lrm) \citep{hong_lrm_2023}.
We first review LRM and its multi-view version.

\paragraph{LRM.} The \lrm \citep{hong_lrm_2023} is a single-view reconstructor.
Given a source image $I_\src$ and its camera $\pi_\src$, \lrm reconstructs a triplane \cite{chan22efficient} representation $\mathcal{T} := \mlrm_\theta(I, \pi)$ of the depicted scene. 
The triplane may be rendered from any target view $\pi_\tgt$ using Emission-Absorption raymarching yielding an RGB render
$R(\pi_\tgt, \mathcal{T})$, depth render $R(\pi_\tgt, \mathcal{T})_D$ and alpha-mask render $R(\pi_\tgt, \mathcal{T})_\alpha$.
In practice, we use the Lightplane renderer \cite{cao_lightplane_2024} to implement $R$.

In a single-view setting, 3D reconstruction is highly ambiguous.
Indeed, as depicted in \cref{fig:inconsistentview}, 
when applied to reconstruct monocular-video frames,
\lrm outputs triplanes with significantly time-inconsistent shape and texture.

\paragraph{Multi-view LRM setup.} Hence, in order to minimize reconstruction ambiguity, we leverage a few-view conditioned version of \lrm. 
Formally, given a set $\mathcal{I}^\src := \{I_\src^i\}_{i=1}^{N^\src}$ of $N^\src$ source images with corresponding cameras
$\Pi^\src := \{\pi_\src^i\}_{i=1}^{N^\src}$
we predict a triplane 
$
\mathcal{T}
:=
\mlrmmv_\theta(\mathcal{I}^\src, {\Pi}^\src)
$, where we overload the same symbol for the multi-view and single-view versions for compactness.
The architecture follows \cite{xu24dmv3d:,li24instant3d:} -- the pixels of each source image $I^\src$ are first concatenated with the Plucker ray coordinates encoding the corresponding camera pose $\pi^\src$ and then fed to DinoV2 \cite{caron21emerging} yielding image tokens.
Then, these tokens enter cross-attention layers inside a large 12-layer transformer that refines a set of fixed shape tokens into the final triplane representation $\mathcal{T}$ of the reconstructed scene.

\paragraph{Multi-view LRM training.}

We train \lrmmv in a fully-supervised manner on a large dataset of artist-created meshes, similar to Objaverse \cite{deitke23objaverse:}. 
We render each mesh from a set of pre-defined camera viewpoints $\Pi$. 
The latter rendering, besides the RGB image $I$, also provides the ground-truth depth map $D$ and the alpha mask $M$.
For each training scene, we sample $N_\src=4$ random images as input views, and render into $N_\tgt=4$ randomly sampled held-out target views where losses are optimized.

We optimize three losses.
(i) The photometric loss 
$
\mathcal{L}_\text{photo}
:= 
\sum_{i=1}^{N_\tgt} 
\| I^i - R(\pi^i, \mathcal{T}) \|^2 
+ \mathrm{LPIPS}(I^i, R(\pi^i, \mathcal{T}))
$;
(ii) mask loss 
$
\mathcal{L}_\text{mask}
:= 
\sum_{i=1}^{N_\tgt} 
\mathrm{BCE}(M^i, R(\pi^i, \mathcal{T})_\alpha)
$, where $\mathrm{BCE}$ is binary cross-entropy;
and (iii) depth loss
$
\mathcal{L}_\text{depth} := \sum_{i=1}^{N_\tgt} \| D^i - R(\pi^i, \mathcal{T})_D \|
$.
Recall that 
$
\mathcal{T}
:=
\mlrmmv_\theta(\mathcal{I}^\src, {\Pi}^\src)
$
is the triplane output by \lrmmv given the 4 source views.
The total loss $\mathcal{L}_\text{photo} + \mathcal{L}_\text{depth} + \mathcal{L}_\text{mask}$ is minimized with the Adam optimizer \cite{KingBa15} with a learning rate of $10^{-4}$ until convergence.

\subsection{\limm: Large Interpolator Model} \label{sec:method_lim}

Given a monocular video, our aim is to predict the 3D representation of the scene at any continuous timestep.
Furthermore, we aim to achieve this in a feed-forward manner, and we require the ability to trace the 3D representation in time, which eventually yields a practically applicable animated mesh with a shared UV texture.

\paragraph{Multi-view \limm.} 
As mentioned in \cref{sec:method_lrm}, reconstructing monocular videos is a highly ambiguous task and, hence, we first focus on the simpler multi-view version with access to multiple views at each timestep.
At the end of this section, we describe how to tackle the harder monocular task by converting it to the multi-view setting described here.

Formally, we are given a multi-view RGB video 
$\{\mathcal{I}_k\}_{k \in (1, 2, \dots, N_f)}$
composed of $N_{f}$ timesteps where, for each integer timestep $k$, we have a set
$\mathcal{I}_k = \{I^i_k\}_{i=1}^{N_v}$ of $N_{v}$ view-points with cameras
$\Pi_k = \{\pi^i_k\}_{i=1}^{N_v}$.
In order to 4D-reconstruct the latter we can, in principle, use \lrmmv to predict a set
$\{\mathcal{T}_k\}_{k \in (1, 2, \dots, N_f)}$
containing a triplane for each keyframe in the video.
However, the latter remains \textbf{discrete} in time and, hence, we cannot obtain a 3D representation at any intermediate continuous timestep $k + \alpha, \alpha \in [0, 1]$.
Furthermore, such frame-specific triplanes encode implicit shapes disconnected across different timesteps. 
This prevents us from converting the time-series of reconstructions into a time-varying mesh.

Thus, to achieve continuous reconstruction in time, and to enable surface tracking, we introduce our Large Interpolator Model (\limm).
Given 2 keyframe sets $\mathcal{I}_k, \mathcal{I}_{k+1}$ at discrete timesteps $k$ and $k+1$, \limm predicts an interpolated triplane $\hat{\mathcal{T}}_{k+\alpha}$ at any continuous timestep $t = k+\alpha, \alpha \in [0, 1]$:
\begin{equation}\label{eq:limm}
    \hat{\mathcal{T}}_{k + \alpha}
    :=
    \mlimm_{\psi}(
        \mathcal{F}_k(\mathcal{I}_k, \Pi_k),
        \mathcal{I}_{k+1},
        \alpha
    ).
\end{equation}
The architecture of \limm, illustrated in \cref{fig:lim_architecture}, takes advantage of the pretrained multiview \lrmmv model from \cref{sec:method_lrm}. 
More specifically, we begin by calculating the intermediate features $\mathcal{F}_k$ as predicted by \lrmmv from the frame set $\mathcal{I}_k$ at the start timestep $k$.
These features are extracted after each of the last $L=6$ transformer blocks of \lrmmv.
Then, we broadcast and concatenate a positional encoding of the interpolation time $\alpha$ to $\mathcal{F}_k$ and feed the result to \limm.
This input is then refined by series of cross-attentions with the image tokens of the next keyframes $\mathcal{I}_{k+1}$ to predict the final interpolated triplane $\hat{\mathcal{T}}_{k + \alpha}$.

\subsection{Training \limm}\label{sec:training_limm}
We train \limm on a large dataset of artist-created meshes animated with a range of motions. 
For each scene, we render the asset from several random viewpoints at each key-frame of the animation. 

In order to train \limm, for each scene, we first sample a pair of keyframe interpolation endpoints at timesteps $k_{\src}$ and $k_{\tgt}$ such that $k_{\tgt} - k_{\src} \in \{2, 3, 4\}$.
Then, we additionally sample a middle keyframe $k_{m}$ such that $k_{\src} \leq k_{m} \leq k_{\tgt}$.
We then task \limm to predict the interpolated triplane
$\hat{\mathcal{T}}_{k_\src + \alpha_m} := \mlimm(\mathcal{F}_{k_\src}, \mathcal{I}_{k_\tgt}, \alpha_m)$
at an intermediate keyframe $k_m$ given the source and target conditioning $\mathcal{F}_{k_{\src}}, \mathcal{I}_{k_{\tgt}}$ and the interpolation time $\alpha_m = \frac{k_{m} - k_{\src}}{k_{\tgt}-k_{\src}}$, which converts the discrete timestep $k_m$ into a continuous interpolation time $\alpha_m \in [0, 1]$.
The interpolated triplane $\hat{\mathcal{T}}_{k_\src + \alpha_m}$ is then compared to the pseudo-ground-truth triplane 
$\mathcal{T}_{k_m} = \mlrmmv(\mathcal{I}_{k_m}, \Pi_{k_m})$
output by \lrmmv at the interpolated keyframe $k_m$ with the following MSE loss:
\begin{equation}\label{eq:basic_lim_loss}
\mathcal{L}_\mathcal{T}
:= 
\| \hat{\mathcal{T}}_{k_\src + \alpha_m}
-
\mathcal{T}_{k_m}
\|^2, 
\alpha_m = \frac{k_{m} - k_{\src}}{k_{\tgt}-k_{\src}}. 
\end{equation}
    
\paragraph{Causal consistency for continuous-time interpolation.}
The loss $\mathcal{L}_\mathcal{T}$ provides a basic supervisory signal which, however, only supervises \limm at keyframe times $k_m$ that are discrete.
The latter prevents the model from becoming a truly temporally-smooth interpolator because, during training, it is never exposed to arbirary interpolation times $\alpha$ spanning the whole continuous range of $[0, 1]$.

To address this, we introduce a causal consistency loss $\mathcal{L}_\text{causal}$.
In a nutshell, the loss enforces that a triplane interpolated directly from time $k_\src$ to $k_\src + \delta, \delta \in [0, 1]$ has to match a triplane that is first interpolated to an arbitrary intermediate timestep $k_\src + \alpha_\text{rand}, \alpha_\text{rand} \in \mathcal{U}(0, \delta)$ and then further interpolated to the target timestep $k_\src + \delta$.

More formally, we define the causal consistency loss as:
\begin{equation}\label{eq:causal_consistency}
\mathcal{L}_\text{causal}
:=
\left\|
\mlimm\left(
    \hat{\mathcal{F}}_{k_\src + \alpha_\text{rand}},
    \mathcal{I}_{k_\src + \delta},
    \frac{\delta - \alpha_\text{rand}}{1 - \alpha_\text{rand}}
\right)
-
\hat{\mathcal{T}}_{k_\src + \delta}
\right\|^2,
\end{equation}
where 
$\hat{\mathcal{F}}_{k_\src + \alpha_\text{rand}}$
stands for the intermediate features predicted by \limm when interpolating from $k_{\src}$ to $k_{\src}+\alpha_\text{rand}$.
Note that we feed into the second \limm pass the intermediate features $\hat{\mathcal{F}}_{k_\src + \alpha_\text{rand}}$ output by \limm as opposed to features $\mathcal{F}$ output by \lrmmv as prescribed by the original \limm formulation in \eqref{eq:limm}.
We empirically observed that this works as we can assume that the intermediate features of \limm follow approximately the same distribution as the intermediate features of \lrmmv.
Hence, \limm can, in a recurrent manner, accept its own intermediate features to ground the interpolation of the next timesteps.
As demonstrated in \cref{sec:exp_interpolation} (\cref{tab:main_4D}), the causal loss $\mathcal{L}_\text{causal}$ significantly improves the temporal consistency and the quality of the interpolations.

Note that for \limm training, \lrmmv model weights $\theta$ are already optimized and we keep them frozen. 
We minimize the total loss $\mathcal{L}_\mathcal{T} + \mathcal{L}_\text{causal}$ using the Adam optimizer \cite{KingBa15} with a learning rate of $10^{-4}$ until convergence.

\paragraph{Monocular 4D reconstruction.}
Given the trained \limm and \lrmmv models, we can now predict the 3D scene representation at any continuous timestep.
However, as mentioned above, the \limm model relies on multiple views at each timestep, which are not always available in practice.

Our method can, however, also be used in the monocular-video to 4D setting.
Here, we leverage a pretrained diffusion model \citep{xie2024sv4ddynamic3dcontent} to recover 3 videos at different viewpoints, consistent in shape and motion with the monocular source.
We then reconstruct 3D per timestep with \lrmmv, and add in-between timesteps (depending on the user frame-rate need), by interpolating with \limm.
This replaces the optimization of a 4D representation from the multi-view videos \cite{bahmani_4d-fy_2023, singer_text--4d_2023}, which in practice takes minutes to hours for a single scene.

\subsection{Tracing shapes with \limm}\label{sec:trace_shapes}

In \cref{sec:training_limm,sec:method_lim}, we have described how \limm, together with \lrmmv, can be trained to predict a continuous-time 3D representation of a scene given a set of multi-view images at discrete timesteps.
As mentioned before, a key goal of our model is to also trace the deformable shape through time.
Next, we describe how to extend \limm and \lrmmv to output canonical surface coordinates to enable surface tracing, and how to use these coordinates to output a time-deforming mesh with fixed topology and texture.

\paragraph{Interpolating canonical surface coordinates.}
To simplify the surface tracing task, we extend \limm and \lrmmv to label the interpolated implicit surface with intrinsic coordinates defined in the canonical coordinate of the object to be interpolated.
More specifically, aside from tasking the interpolated triplanes $\mathcal{T}$ with representing the RGB color and geometry of the implicit shape, we also task them with supporting a volumetric function $f: \mathbb{R}^3 \rightarrow \mathbb{R}^3$ that maps each point in the 3D space to its canonical surface coordinate.
Without loss of generality, we set the canonical coordinates of time-deforming shape to the XYZ coordinates of the corresponding surface points in the start timestep $k_\src$.
Since we have a dataset of artist-created meshes with a known deformation of each vertex in time, we can easily calculate the canonical coordinate function $f_{k_\src}$ of each vertex at the start timestep $k_\src$ and then transport those using the known animation to any other deformation time $k_\tgt$ yielding a function $f_{k_\tgt}$.
Importantly, $f$ can be rendered at any point in time from a viewpoint $\pi^i$ yielding a 3-channel canonical coordinate render $C^i$.

Given the above, we train a second \lrmmv, dubbed $\overline{\mlrmmv}$, which shares the same architecture, but predicts triplanes $\mathcal{T}^C := \overline{\mlrmmv}(\mathcal{C}_\src, \Pi_\src)$ supporting the coordinate function $f$, by accepting a set $\mathcal{C}_\src := \{C^i_\src\}_{i=1}^{N_\src}$ of multi-view source canonical renders $C^i_\src$.
This canonical-coordinate $\overline{\mlrmmv}$ is supervised with the following canonical loss:
\begin{equation}
    \mathcal{L}_\text{can} := \| C^i_\src - R(\pi^i, \mathcal{T}_\src^C) \|^2,
\end{equation}
and with the same depth, and mask losses as the \lrmmv in \cref{sec:method_lrm}.
Similarly, we train $\overline{\mlimm}$ to interpolate the canonical-coordinate triplane as, 
\begin{equation}
\hat{\mathcal{T}}_{\src}^C := \overline{\mlimm}(\mathcal{F}_{k_\src}^C, \mathcal{I}_{k_\src}, \mathcal{I}_{k_\tgt}, \alpha).
\end{equation}
Note that this $\overline{\mlimm}$, analogous to \limm, is conditioned on the features $\mathcal{F}_{k_\src}^C$ of the canonical-coordinate $\overline{\mlrmmv}$.
However, differently from \limm, $\overline{\mlimm}$ accepts target and source RGB frames $\mathcal{I}_{k_\src}$ and $\mathcal{I}_{k_\tgt}$ instead of the target-time canonical image $\mathcal{C}_{k_\tgt}$.
This is because the canonical coordinates can only be carried forward in time and, as such, are not available for the target timestep $k_\tgt$.
Hence, $\overline{\mlimm}$ instead learns how to propagate the coordinates by
analyzing the RGB frames that are available in both timesteps. 
We supervise $\overline{\mlimm}$ with the MSE loss $\mathcal{L}_\mathcal{T}^C$ and the causal consistency loss $\mathcal{L}_\text{causal}^C$ that are defined analogously to \eqref{eq:basic_lim_loss} and \eqref{eq:causal_consistency} but with the canonical-coordinate triplanes $\mathcal{T}^C$ and images $\mathcal{C}$.

\paragraph{Mesh tracing.}
Given the RGB and canonical-coordinate versions of \limm and \lrmmv, we can trace a mesh multi-view frames $\mathcal{I}_\src$ and $\mathcal{I}_\tgt$ (recall that, using an image diffusion model, this is also possible for a monocular video).

We start by extracting the color triplane $\mathcal{T}_{k_\src}$ at timestep $k_\src$ with \lrmmv.
We then render $\mathcal{T}_{k_\src}$ to obtain a depth map $D_{k_\src}$ that we unproject to form 3D points yielding the multi-view canonical coordinates $\mathcal{C}_{k_\src}$.
Given $\mathcal{C}_{k_\src}$, we can predict the canonical-coordinate triplane $\mathcal{T}_{k_\src}^C$ with $\overline{\mlrmmv}$.

Then, for a series of monotonic time offsets 
$\alpha_0, \dots, \alpha_N; |\alpha_{j+1}-\alpha_j| \rightarrow 0$ 
the canonical coordinate triplane $\mathcal{T}_{k_\src}^C$, 
together with $\mathcal{I}_\src$ and $\mathcal{I}_\tgt$, is fed to $\overline{\mlimm}$ to interpolate the canonical-coordinate triplane $\hat{\mathcal{T}}_{k_\src + \alpha_j}^C$ at all continuous timesteps $k_\src + \alpha_j$.

The series of resulting canonical-coordinate triplanes $\hat{\mathcal{T}}_{k_\src + \alpha_j}^C$ provides a series of implicit shapes annotated with surface coordinates.
To obtain a time deforming mesh, we first run Marching Cubes (MC) \cite{lorensen87marching} on the first triplane $\hat{\mathcal{T}}_{k_\src + \alpha_0}^C$ resulting in a mesh $\mathcal{M}_{k_\src + \alpha_0}(V_{k_\src + \alpha_0}, F)$ with time-dependent vertices $V_{k_\src + \alpha}$ and time-invariant faces $F$.
We then run MC on the next triplane $\hat{\mathcal{T}}_{k_\src + \alpha_1}^C$ and match the vertices of the previous mesh to the surface on the next mesh using nearest neighbor search in the space of canonical coordinates defined by the triplanes $\hat{\mathcal{T}}_{k_\src + \alpha_0}^C$ and $\hat{\mathcal{T}}_{k_\src + \alpha_1}^C$, respectively.
Afterwards, we replace the vertices $V_{k_\src + \alpha_0}$ with the corresponding nearest neighbors from the next time $k_\src + \alpha_1$, and repeat the process for all the remaining timesteps $\alpha_2, \dots, \alpha_N$

\section{Experiments}
\label{sec:experiments}

\begin{figure*}
  \centering
  \includegraphics[width=\textwidth]{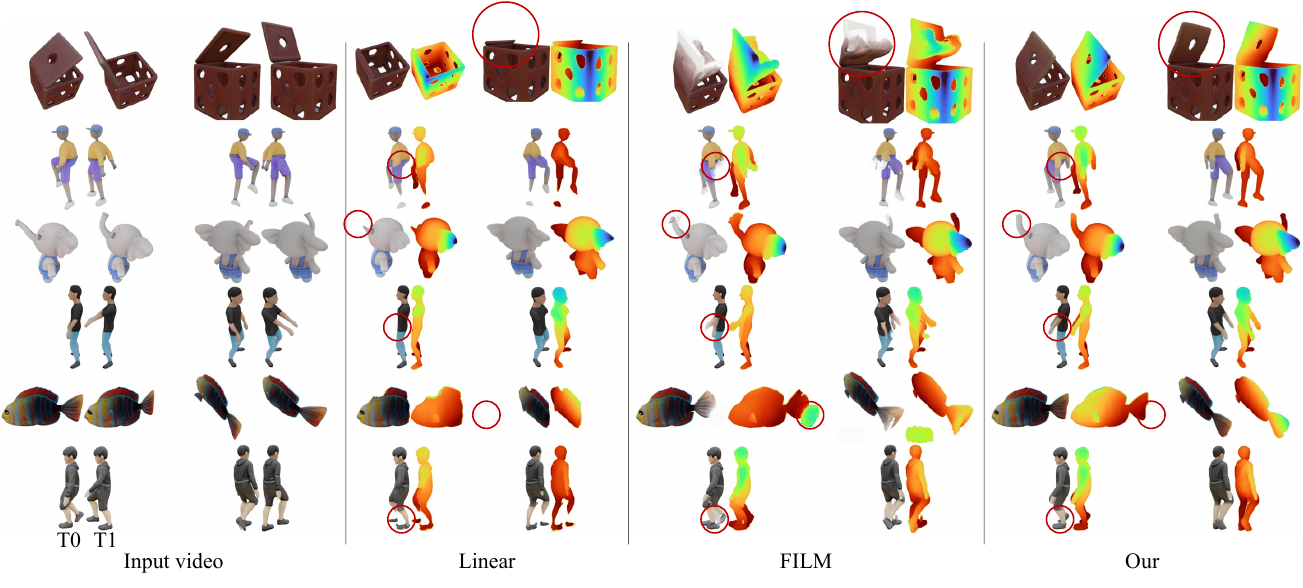}
   \caption{\textbf{Interpolation results} comparing 
   (i)~linear interpolation in triplane space, which fails on dynamic parts;
   (ii)~image-based interpolator \citep{reda2022filmframeinterpolationlarge} (FILM), yielding view-consistent frame interpolations leading to defective reconstructions (ghosting around dynamic parts; for example, the tip of the elephant's trunk or fish's tail); and 
   (iii)~our \limm-based interpolation, which yields the most plausible results.
   }
   \label{fig:qualitative_interpolation}
   \vspace{-0.1in}
\end{figure*}

\begin{table}[b!]
\centering
\scriptsize
\caption{\textbf{Interpolation results} comparing \limm to a linear triplane interpolation, and to an image-based interpolation implemented with the FiLM image interpolator \cite{reda2022filmframeinterpolationlarge}.
\label{tab:main_interpolation}}
\begin{tabular}{r|ccc}
\toprule
 & PSNR $\uparrow$  & $\text{PSNR}_\text{FG} \uparrow$  & LPIPS $\downarrow$ \\ \midrule
Linear & 20.96 & 11.04 & 0.093 \\
FILM \citep{reda2022filmframeinterpolationlarge} & 22.05 & 14.98 & 0.082 \\
\limm (Our)  & 23.11 & 16.12 & 0.075 \\
\midrule
Oracle & 24.43 & 17.51 & 0.064 \\
\bottomrule
\end{tabular}
\end{table}

\subsection{Feed-forward Triplane Interpolation} \label{sec:exp_interpolation}

In this section, we evaluate the ability of \limm to interpolate triplanes so that the renders of the latter match the ground-truth views extracted at the target interpolation timestep.
More specifically, given a multi-view video of a deforming object, which contains the frame set $\{\mathcal{I}_k\}_{k=1}^{N_f}$ at each timestep $k$, we first split the frame sets into adjacent triplets $\mathcal{I}_k, \mathcal{I}_{k+1}, \mathcal{I}_{k+2}$ for a $k$-th triplet.
In each triplet, we then evaluate the ability of a method to interpolate the 3D representation $\hat{\mathcal{T}}_{k+1}$ at the mid-point $k+1$ given the boundary frames $\mathcal{I}_k, \mathcal{I}_{k+2}$.
Note that, since the frames are sampled at uniform time intervals, the continuous interpolation index is $\alpha = 0.5$.
For evaluation, we select 256 scenes from our heldout dataset of animated objects.

Given the interpolant $\hat{\mathcal{T}}_{k+1}$, we then evaluate its quality by rendering into the set of (novel) evaluation views, and reporting three photometric errors measuring the discrepancy between the renders and the corresponding ground-truth images: (i) peak signal-to-noise ratio~\textbf{PSNR};  (ii)~perceptual loss \textbf{LPIPS}~\cite{zhang2018perceptual}; and (iii)~$\textbf{\text{PSNR}}_\text{\textbf{FG}}$ calculating PSNR only over the foreground pixels.

\paragraph{Baselines.} We compare our \limm interpolation with two baselines. 
The first baseline (Linear) is a simple linear interpolation, which defines the interpolated triplane
$\hat{\mathcal{T}}_{k+1}^\text{linear} = (1-\alpha) \mathcal{T}_{k} + \alpha \mathcal{T}_{k+2}$ as a linear combination of the two triplanes predicted by \lrmmv for each set of boundary frames.
The second baseline (FILM) is image-based.
Specifically, we first interpolate in the image-space using a pre-trained deep image interpolator \emph{FILM} \citep{reda2022filmframeinterpolationlarge} which, given the boundary views $I^i_k, I^i_{k+2}$, generates the interpolant $\hat{I}^i_{k+1}$ followed by multi-view \lrmmv reconstruction yielding the interpolated triplane $\hat{\mathcal{T}}_{k+1}^\text{FILM}$.
We also report results from an Oracle method, which has access to the ground-truth images $\mathcal{I}_{k+1}$ and reconstructs the corresponding triplane $\hat{\mathcal{T}}_{k+1}^\text{Oracle}$ with \lrmmv.
The latter provides an upper performance limit.

\begin{figure*}
  \centering
  \includegraphics[width=\textwidth]{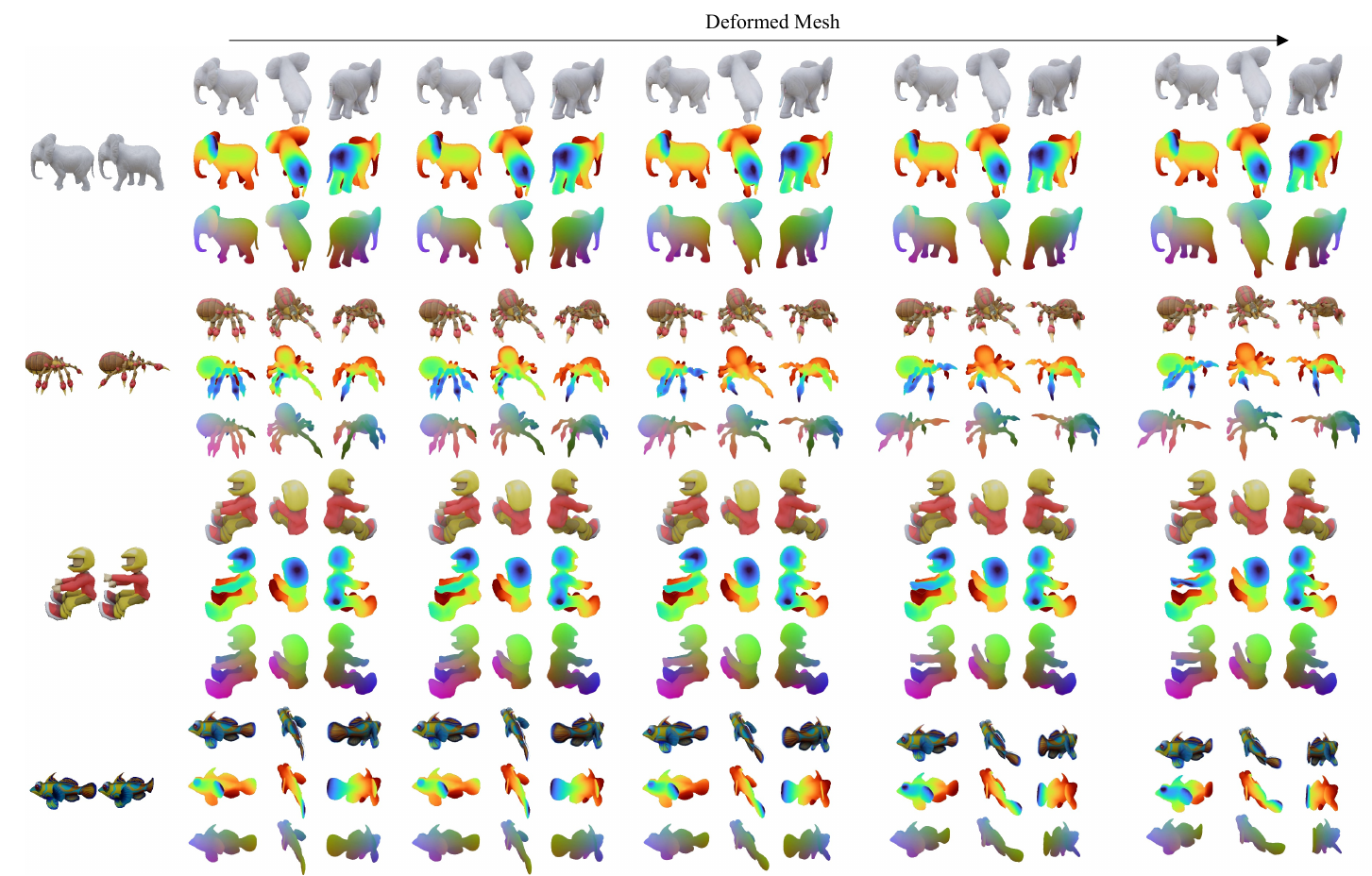 }
   \caption{\textbf{Mesh Tracking results.} Given two implicit 3D representations, \limm can interpolate densely in time and hence can track a source mesh to produce a deforming mesh sequence. For each scene, we show (top to bottom) RGB rendering of the tracked mesh, depth and canonical-coordinate interpolation. See supplemental video on the \webpagelink{project page}.}
   \vspace{0.1in}
   \label{fig:qualitative_meshtrack}
\end{figure*}

\paragraph{Results.} \cref{tab:main_interpolation} presents the results.
We also provide \cref{fig:qualitative_interpolation} and a supplemental video for visual evaluation.
\limm outperforms linear interpolation and image-based interpolation on all three metrics.
Here, we notice that linear interpolation in the triplane space fails to correctly represent dynamic elements, which often disappear after being interpolated.
Furthermore, the image-based interpolation often results in artifacts in the color and opacity fields, which is due to view-inconsistencies between the images interpolated at the same timestep. 
\limm scores closest to the Oracle bound.

\paragraph{Ablating causal consistency.}
\Cref{tab:main_ablation} reports the performance of a \limm trained without the causal consistency loss $\mathcal{L}_\text{causal}$ on the aforemention benchmark.
The evaluation reveals a significant drop in performance, confirming the merit of the causal loss.
See supplementary material.

\begin{table}[h!]
\centering
\caption{
\textbf{Ablating $\mathcal{L}_\text{causal}$.}
Interpolation accuracy comparing our \limm with its ablation removing the causal consistency loss $\mathcal{L}_\text{causal}$.
\label{tab:main_ablation}
}
\begin{tabular}{r|ccc}
\toprule
& PSNR $\uparrow$  & $\text{PSNR}_\text{FG} \uparrow$  & LPIPS $\downarrow$ \\ \midrule
\limm - wo/ $\mathcal{L}_\text{causal}$ & 22.2 & 15.38 & 0.084 \\
\limm & \textbf{23.11} & \textbf{16.12} & \textbf{0.075} \\
\bottomrule
\end{tabular}
\end{table}

\begin{table}[b!]
\centering
\caption{\textbf{Evaluation of deformable mesh tracking} comparing our \limm with Nearest-Neighbor tracing}
\begin{tabular}{r|ccc}
\toprule
 & PSNR $\uparrow$  & $\text{PSNR}_\text{FG} \uparrow$  & LPIPS $\downarrow$ \\ \midrule
NN-tracing & 20.33 & 16.09 & 0.122 \\
LIM (Our) & \textbf{21.56} & \textbf{17.11} & \textbf{0.096} \\
\bottomrule
\end{tabular}
\label{tab:main_tracing}
\end{table}

\subsection{Deformable Mesh Reconstruction}

In this section, we evaluate the quality of the dynamic mesh reconstructions output by \limm's mesh-tracing method (\cref{sec:trace_shapes}).
More specifically, we create a dataset of 8-step test sequences heldout from the train set.
As before, the $k$-th timestep of the 8 timesteps contains a frame set $\mathcal{I}_k$.
First, we reconstruct a mesh in the canonical pose, defined as the shape at the first frame ($k=1$).
We then use our mesh-tracing method to deform the vertices of the mesh so they follow the motion observed in the frame sets $\mathcal{I}_k, k \in [2, \dots, 8]$.
Note that we keep the topology of the first-frame mesh, as well as its texture shared across all 8 timesteps.
After the mesh is traced, we render it at timesteps $\{4,6,8\}$ to several heldout views and again evaluate the PSNR, LPIPS, and $\text{PSNR}_\text{FG}$. 
We compare our \limm mesh tracing, described in \cref{sec:trace_shapes}, with a baseline (Nearest Matching) that iteratively deform the vertices of the mesh at timestep $i$ to the nearest match on the surface of the mesh at timestep $i+1$, where the matching is performed over distance and RGB-features.

\begin{figure*}[t!]
  \centering
  \includegraphics[width=.9\textwidth]{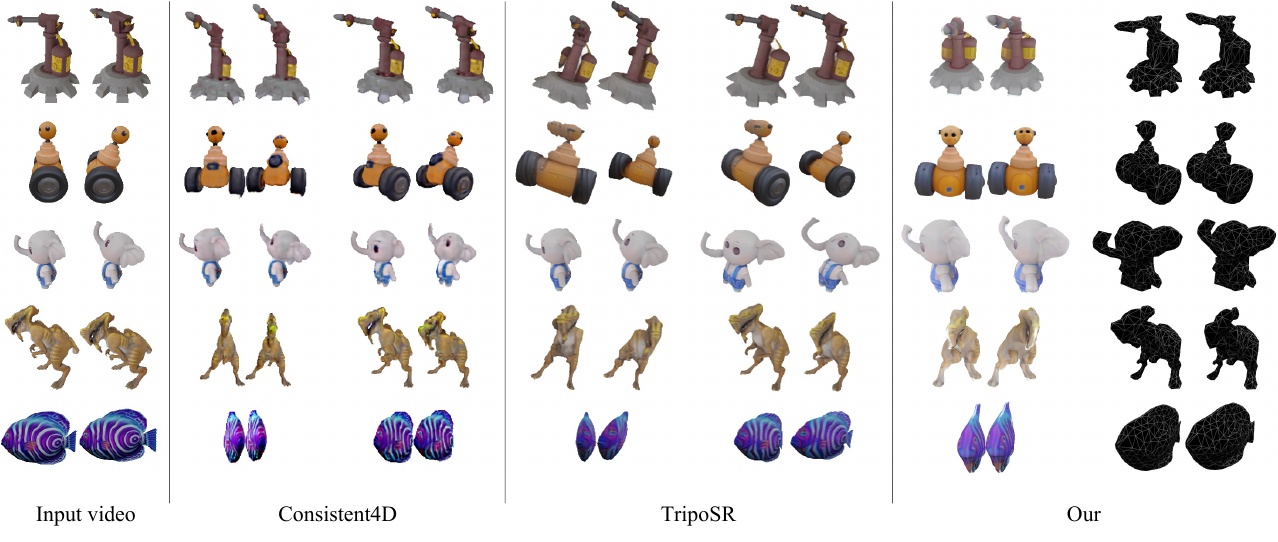}
   \caption{\textbf{Monocular 4D Reconstruction}
   comparing \limm with Consistent4D and TripoSR applied to each input frame separately. Our method is the only one to output a time-deforming mesh with \text{fixed} topology and texture. For our method, we render the topology for the second view.
   }
   \label{fig:qualitative_monocular}
\end{figure*}

\paragraph{Results.} 
\cref{tab:main_tracing} present the quantitative results.
Our main observation is that LIM's ability to densely and accurately interpolate RGB and XYZ values over time, allows one to avoid explicitly solving the challenging  correspondence problem between keyframed poses, separated by non-trivial deformations. 
Our evaluations show that LIM works well overall, although the results degrade around thin structures. 
Additional results are available at the \webpagelink{project page}.

\subsection{4D Reconstruction} \label{sec:exp_4D}

Finally, we evaluate \limm on 4D reconstruction from a monocular video.

For each evaluation scene, we extract a video sequence of 16 frames $\{I^1_{k} \}_{k=1}^{16}$.
We then leverage an external diffusion-based model to generate 3 additional views of the scene $\{ \hat{I}^i_{k}\}_{k \in [1,\dots,16], i \in \{2,3,4\}}$ for each timestep. 
We then reconstruct the 3D representation on the odd frames with \lrmmv and interpolate with our \limm to predict the representation on the even frames.
We evaluate on all the even frames, on a set of four random views, outside the training views.
We report two metrics:  the LPIPS error between the ground-truth images and the renders of the reconstructed triplanes, and the FVD \cite{Unterthiner2019FVDAN} measuring comparing generated new-view sequence against ground-truth temporal sequence of renders.

\paragraph{Baselines.} We compare two baselines: (i) Consistent4D~\citep{jiang_consistent4d_2023}, an optimization-based model, conditioned on monocular video and supervised via SDS loss;
(ii) TripoSR~\citep{tochilkin_triposr_2024}, an open-source LRM conditioned on a single image, \ie, the model is at a disadvantage.

\paragraph{Results.} Evaluation results are in \cref{tab:main_4D}.
The combination of \limm with multi-view diffusion model outperforms the competing methods by a significant margin.

\begin{table}[h!]
\centering
\scriptsize
\caption{\textbf{Monocular video reconstruction} results comparing our \limm to Consistent4D \cite{jiang_consistent4d_2023} and to TripoSR \citep{tochilkin_triposr_2024} applied independently to each frame of the input video.
\label{tab:main_4D}}
\begin{tabular}{r|cc|cc}
\toprule
& Feed-fwd. & Inf. Time & LPIPS $\downarrow$ & FVD $\downarrow$ \\ \midrule
Consistent4D \citep{jiang_consistent4d_2023}  & \faTimes & $\sim$1.5hours 
& 0.429 & 1136.3 \\
TripoSR \citep{tochilkin_triposr_2024} & \faCheck & $\sim$30secs 
& 0.504 & 1427.2 \\
\limm (Ours) & \faCheck & $\sim$3min 
& \textbf{0.142} & \textbf{811.1} \\
\bottomrule
\end{tabular}
\end{table}

\section{Conclusion}
\label{sec:conclusion}

We have proposed \limm, a novel method paired with multiview \lrmmv to enable continuous and feed-forward rendering in both space and time. As opposed to image-based interpolators or direct triplane baselines, we demonstrated  that \limm results in high-quality and consistent 4D interpolations, realistically capturing deformations, and can support different type of modalities. A key advantage of ours is that we can interpolate in RGB and (canonical) XYZ, which in turn allows to directly output consistently-textured dynamic mesh assets that applicable in production workflows. 

A key limitation of our approach is it being trained on synthetic data. Since \limm learns to interpolate  deformation/motion, rather than appearance, we expect the results to carry over to real data. However, we would need an LRM model trained on real-world data to test this hypothesis. 
We leave this for later exploration. Also, in the future, we would like to extend our framework to handle extrapolation, instead of interpolation. The challenge would be to effectively use video data and video generators for training.

\section*{Acknowledgment}
The work was partially supported by the UCL AI Centre and a UCL-Meta PhD support. We thank Tom Monnier for his valuable guidance during the rebuttal stage.

{
    \small
\bibliographystyle{ieeenat_fullname}

}
\clearpage
\appendix
\setcounter{page}{1}

\twocolumn[{
\centering
\Large
\textbf{\thetitle}\\
\vspace{0.5em}Supplementary Material \\
\vspace{1.0em}
\begin{center}
\includegraphics[width=\textwidth]{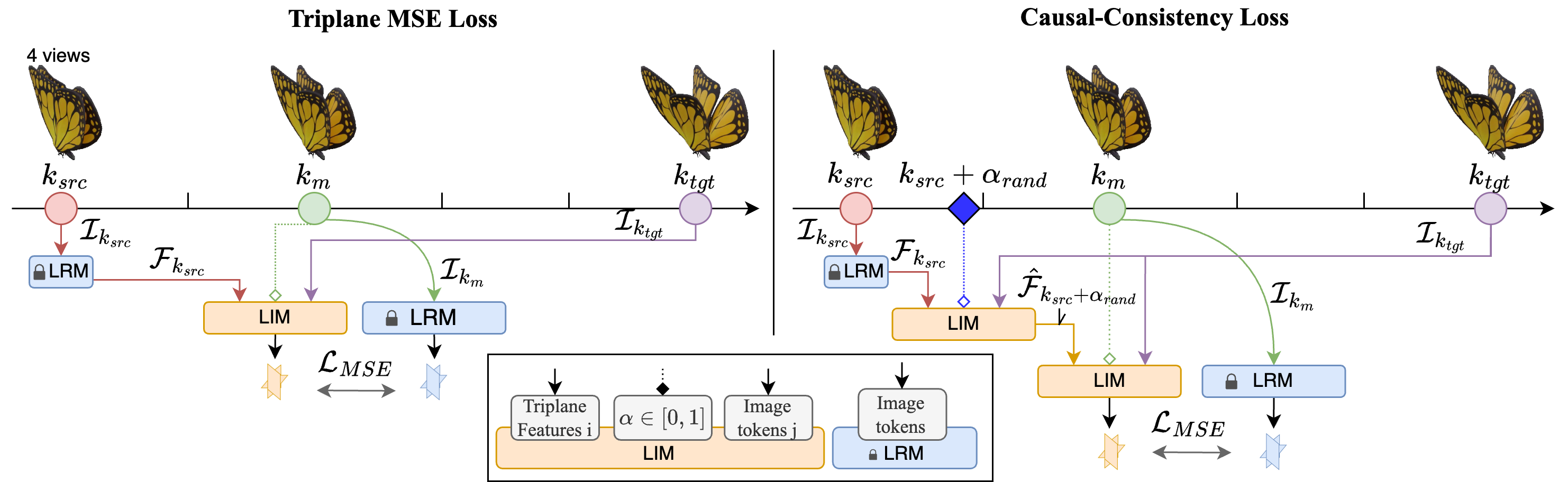}
\end{center}
\vspace{-0.5cm}
\captionsetup{type=figure}
\captionof{figure}{%
\textbf{LIM training losses}.
(Left) The triplane MSE loss $\mathcal{L}_\mathcal{T}$ only supervises \limm on keyframes $k_{m}$. (Right) The causal consistency loss  $\mathcal{L}_\text{causal}$ samples in-between keyframes with an additional forward-pass to \limm.
Note that the second pass of \limm takes as input the intermediate features from \limm instead of the intermediate features from \lrmmv.
}
\label{fig:lim_causal_loss}
\vspace{0.7cm}
}]

\section{Additional Evaluations}
We recommend looking at the \webpagelinkflat{project page}, to see the video results.
In particular, the webpage contains video result of RGB interpolation, XYZ canonical tracking, monocular reconstruction and mesh reconstruction.

\paragraph{Evaluation on OOD data}
We provide qualitative results on the Consistent4D eval set, which includes \textit{real-world scenes}, in Table \ref{tab:reb_monocular}.

\section{Additional Method Insights}

\begin{figure*}[h!]
\centering
\includegraphics[width=\textwidth]{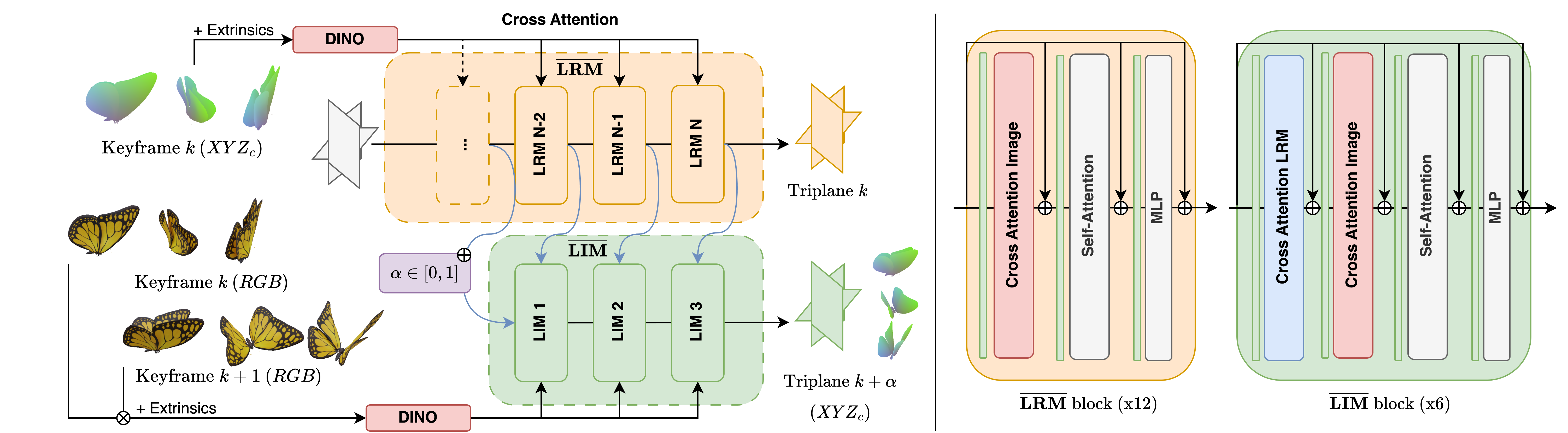}
\caption{\textbf{$\overline{\mlimm}$ framework}. (Left) Given multi-view RGB images on 2 timesteps $k$ and $k+1$ and XYZ canonical renders on timestep $k$, $\overline{\mlimm}$ interpolates any intermediate 3D representation of the XYZ canonical coordinate at $k+\alpha, \alpha \in [0,1]$. This gives direct correspondences in 3D space between the source shape at $k$ and the interpolated shape at $k+\alpha$. In practice, our $\overline{\mlimm}$ architecture has 6 blocks and $\overline{\mlrmmv}$ 12 blocks. (Right) Block structure of $\overline{\mlrmmv}$ and $\overline{\mlimm}$. We include layer normalization before each module in blocks.}
\label{fig:lim_architecture_xyz}
\end{figure*}

\paragraph{Weight Initialization.}
The composition of blocks in \limm and \lrmmv is presented in \cref{fig:lim_architecture}.
We initialize \limm with \lrmmv to take advantage of the learned 3D intermediate representation.
More specifically, the intermediate-features cross-attention layers are derived from the self-attention layers from \lrmmv. 
Furthermore, the image cross-attention layers are initialized using the image cross-attention layers from \lrmmv, and the self-attention layers are initialized from the self-attention layer of \lrmmv. Initialization is similar for $\overline{\mlrmmv}$ and $\overline{\mlimm}$ (presented in \cref{fig:lim_architecture_xyz}).

\paragraph{Model size.}
We ablate the choice of the number of layers in \cref{tab:ablation_layer}.
We observe that \limm accuracy is proportional to the number of blocks in the architecture.
However, adding more blocks in \limm slows down the interpolation.
We set $N_{layer} = 6$ as a good trade-off between speed and accuracy. 

\paragraph{Dataset details}
Our \emph{3D} dataset includes 142,123 assets, while the \emph{4D} dataset comprises 6,052 rigged models, each with 16 to 128 keyframes. We render the keyframes using Blender and the Cycles engine.

\begin{figure}[b!]
\centering
\includegraphics[width=\columnwidth]{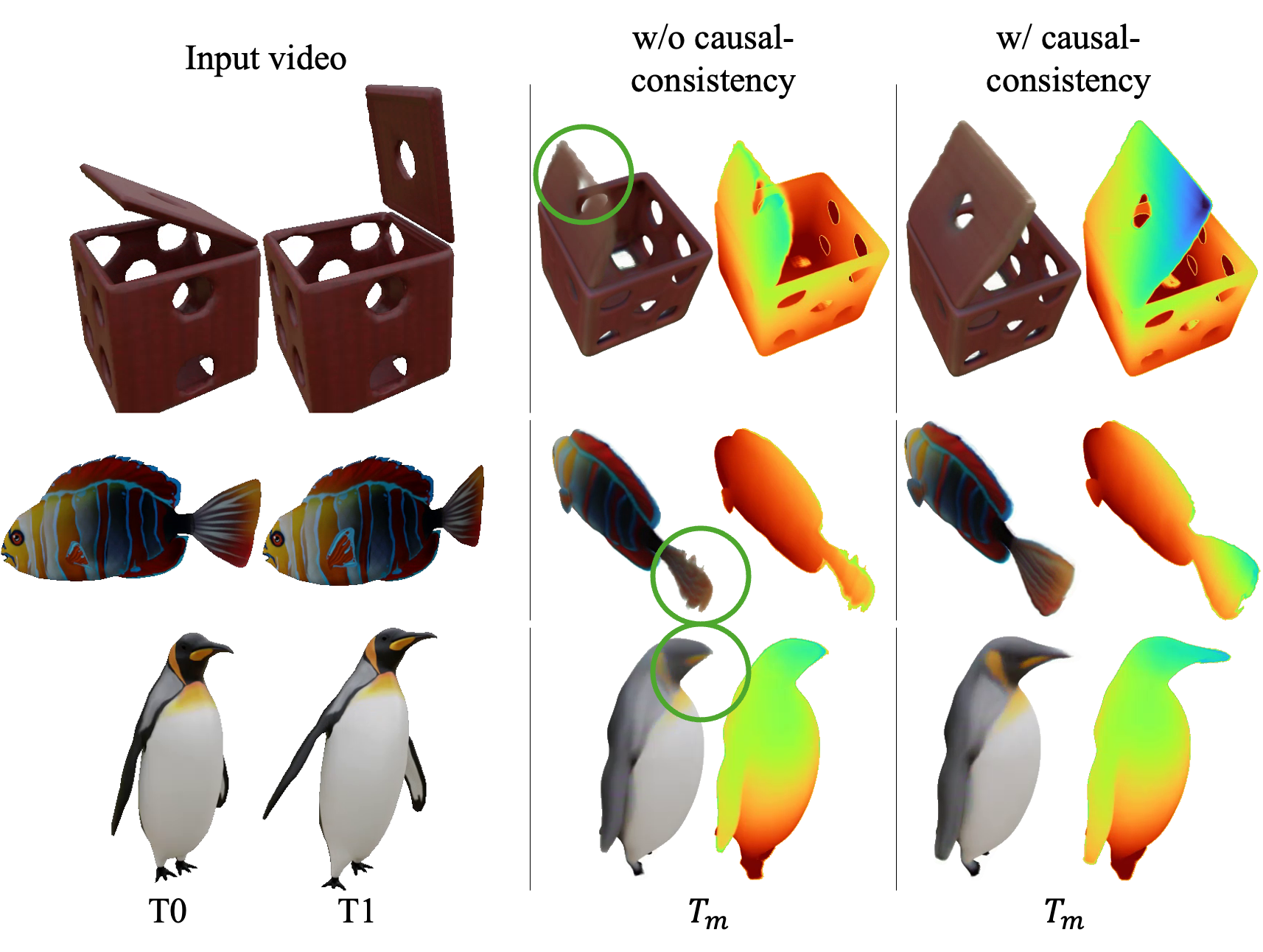}
\caption{\textbf{Causal-loss ablation.}
We show triplane interpolation result from \limm models trained either with the triplane MSE loss $\mathcal{L}_\mathcal{T}$ only, or with both $\mathcal{L}_\mathcal{T}$ and the causal-consistency loss $\mathcal{L}_\text{causal}$.}
\label{fig:qualitative_causal}
\end{figure}

\begin{table}[b!]
\centering
\scriptsize
\caption{\textbf{Monocular reconstruction (out of distribution OOD)}.}
\label{tab:reb_monocular}
\vspace{-0.1in}
\begin{tabular}{r|c|cc}
\toprule
 & Inf. Time & \multicolumn{2}{c}{Consistent4D set} \\ 
\cmidrule(lr){3-4}
 & & LPIPS & FVD \\ 
\midrule
Consistent4D& $\sim$90 min & 0.428 & 1134.7 \\
TripoSR     & $\sim$0.5 min & 0.497 & 1428.2 \\
\limm (Our) & $\sim$3 min  & \textbf{0.114} & \textbf{781.9} \\ 
\bottomrule
\end{tabular}
\end{table}

\begin{table}[h]
\centering
\begin{tabular}{r|ccc}
\toprule
& PSNR $\uparrow$  & $\text{PSNR}_\text{FG} \uparrow$  & LPIPS $\downarrow$ \\ \midrule
\limm - 3 layers & 22.35 & 14.56 & 0.079 \\
\limm - 8 layers & 23.19 & 16.2 & 0.075 \\
\limm & 23.11 & 16.12 & 0.075 \\
\bottomrule
\end{tabular}
\caption{%
\textbf{Performance as a function of \# layers}
reporting interpolation accuracy of \limm while varying the number of transformer blocks in the architecture.
\label{tab:ablation_layer}
\vspace{-0.3cm}
}
\end{table}

\paragraph{Causal consistency loss.}
We illustrate in \cref{fig:lim_causal_loss} the behavior of the triplane MSE loss $\mathcal{L}_\mathcal{T}$ and the causal-consistency loss $\mathcal{L}_\text{causal}$ (see \cref{sec:method}).
$\mathcal{L}_\mathcal{T}$ involves a single pass of \limm and two passes of \lrmmv, while $\mathcal{L}_\text{causal}$ involves 2 passes of \lrmmv and 2 passes of \limm.
Note that during \limm training, the weights of \lrmmv are frozen.
In practice, we discovered that the causal consistency loss was essential to achieve precise and accurate interpolation over a range of shapes and motions.
We show interpolation results (in the same setting as \cref{sec:exp_interpolation}) in \cref{fig:qualitative_causal}, with a \limm model trained either with $\mathcal{L}_\text{causal}$ activated or deactivated.

\paragraph{Positional Encoding}
We apply positional encoding to the interpolation time $\alpha \in [0,1]$ with $\phi: \mathbb{R} \rightarrow \mathbb{R}^{2D}$, such that 
$\forall i \in [1,D], \phi(\alpha)[2i] = \cos(\alpha f_{2i}); \phi(\alpha)[2i+1] = \sin(\alpha f_{2i+1})$, and $f_i = \exp[{-\frac{\log{10.000}}{D}.i}]$; we set $D=512$ so that $2D$ matches the LRM embedding dimension.

\paragraph{4D reconstruction with ARAP regularization}.
We observe that our mesh-tracking framework can incorporate ARAP regularization to mitigate issues like triangle inversion or self-intersection. 
Instead of relying solely on direct matching through nearest neighbor search in the space of canonical coordinates (refer to Section \cref{sec:trace_shapes}), we implement a concise optimization loop. This loop incorporates both canonical-coordinate matching and ARAP energy as objectives to minimize.

\end{document}